\documentclass[10pt,twocolumn,letterpaper]{article}

\usepackage[pagenumbers]{cvpr} % To force page numbers, e.g. for an arXiv version

\usepackage{url}
\usepackage{amssymb}
\usepackage{booktabs} % For toprule, midrule, and bottomrule
\usepackage{multirow} % For multi-row cells
\usepackage{colortbl} % For cell coloring
\usepackage{xcolor}   % For custom colors
\usepackage{pifont}   % For special symbols (e.g., diamonds)
\usepackage{graphicx} % For including images
\usepackage{tabularx} % For automatic column width adjustment
\usepackage{pifont} % For uniform checkmarks and crosses
\newcommand{\cmark}{\ding{51}} % Checkmark symbol
\newcommand{\xmark}{\ding{55}} % Cross symbol
\usepackage{makecell}

\usepackage[most]{tcolorbox} 
\tcbset{
    colback=gray!20,     % 背景颜色
    arc=5mm,               % 圆角半径
    boxrule=0.3mm,         % 边框粗细
    width=\linewidth,      % 让盒子充满当前行宽
    left=2mm,              % 内左边距
    right=2mm,             % 内右边距
    top=2mm,               % 内上边距
    bottom=2mm,            % 内下边距
}

\usepackage{listings}
\usepackage{color}
\definecolor{codegreen}{rgb}{0,0.6,0}
\definecolor{codegray}{rgb}{0.5,0.5,0.5}
\definecolor{codepurple}{rgb}{0.58,0,0.82}
\definecolor{backcolour}{rgb}{0.95,0.95,0.92}
\lstdefinestyle{mystyle}{
    backgroundcolor=\color{backcolour},   
    commentstyle=\color{codegreen},
    keywordstyle=\color{magenta},
    numberstyle=\tiny\color{codegray},
    stringstyle=\color{codepurple},
    basicstyle=\ttfamily\scriptsize,
    breakatwhitespace=false,         
    breaklines=true,                 
    captionpos=b,                    
    keepspaces=true,                 
    numbers=left,                    
    numbersep=5pt,                  
    showspaces=false,                
    showstringspaces=false,
    showtabs=false,                  
    tabsize=2,
    showlines=true
}

\newcommand\blfootnote[1]{% 
    \begingroup 
    \renewcommand\thefootnote{}\footnote{#1}% 
    \addtocounter{footnote}{-1}% 
    \endgroup 
}
\usepackage[hang,flushmargin]{footmisc}

\definecolor{cvprblue}{rgb}{0.21,0.49,0.74}
\usepackage[pagebackref,breaklinks,colorlinks,allcolors=cvprblue]{hyperref}

\def\paperID{***} % *** Enter the Paper ID here
\def\confName{CVPR}
\def\confYear{2025}

\title{Enhancing Video-LLM Reasoning via Agent-of-Thoughts Distillation}

\vspace{-2em}
\author{Yudi Shi$^{1,2}$, Shangzhe Di$^{1,2}$, Qirui Chen$^{1,2}$, Weidi Xie$^{1,\dagger}$ \\[3pt]
$^{1}$School of Artificial Intelligence, Shanghai Jiao Tong University, China\\[2pt]
$^{2}$Coop. Medianet Innovation Center, Shanghai Jiao Tong University, China\\[2pt] \\[-12pt]
\textbf{\url{https://zhengrongz.github.io/AoTD/}}
}

\begin{document}
\maketitle

\blfootnote{
$\dagger$: Corresponding author.
}

\begin{figure*}[t]
  \centering
  \resizebox{.99\linewidth}{!}{
  \includegraphics{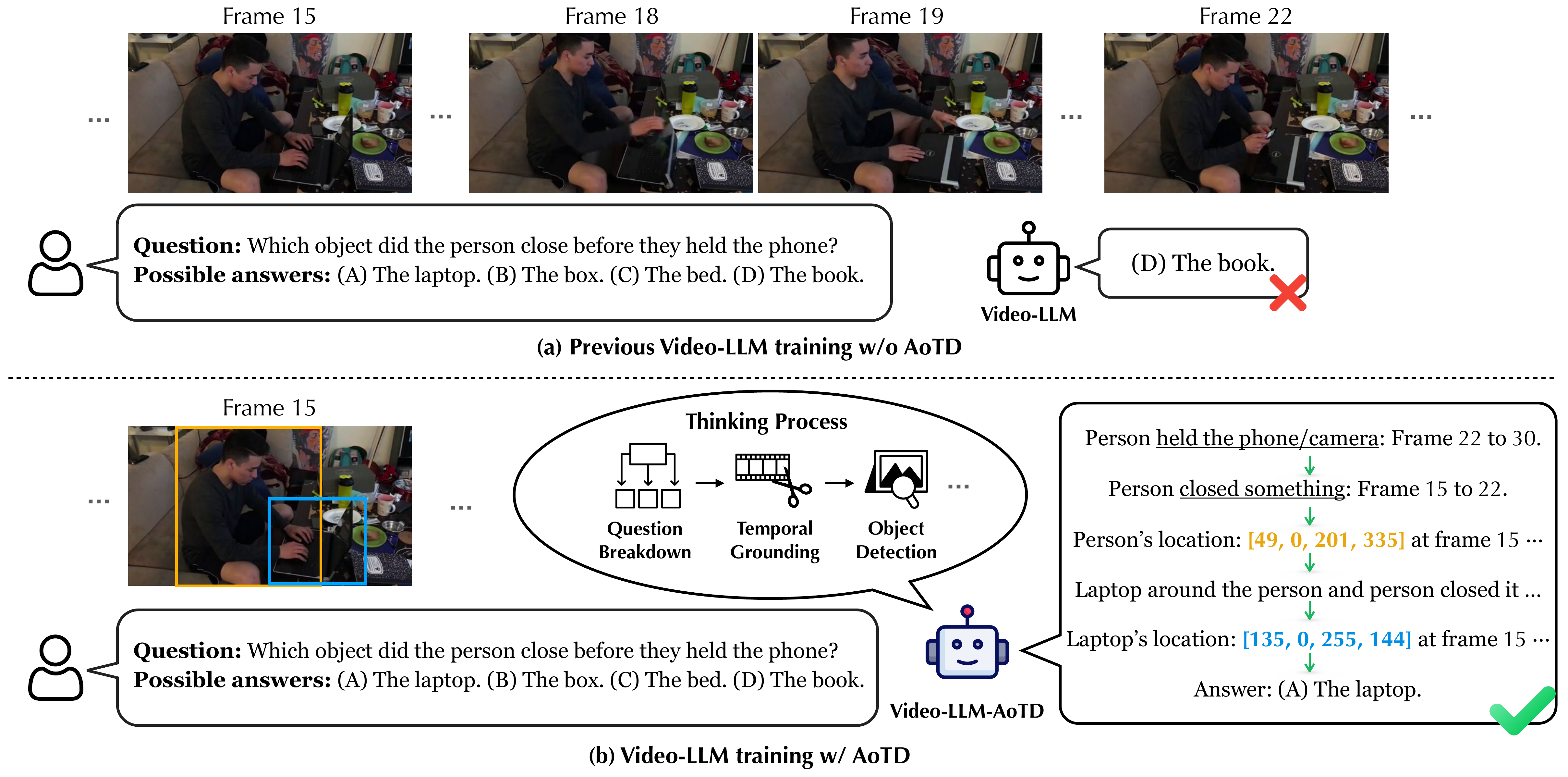}
  }
  \vspace{-5pt}
  \caption{
  Our method, \textbf{AoTD}, distills multi-step reasoning and spatial-temporal understanding into a single generative video-language model. 
  When addressing complex VideoQA tasks, the model trained with AoTD (as shown in (b)) enables to generate a step-by-step reasoning to get the correct answer. In contrast, previous models trained solely on question-answer pairs (as in (a)) generate only a final answer, often without intermediate reasoning, which can lead to incorrect conclusions.
  \vspace{-1.1em}
  }
  \label{fig: teaser}
\end{figure*}

\begin{abstract}

This paper tackles the problem of video question answering (VideoQA), 
a task that often requires multi-step reasoning and a profound understanding of spatial-temporal dynamics. While large video-language models perform well on benchmarks, they often lack explainability and spatial-temporal grounding. 
In this paper, we propose \textbf{A}gent-\textbf{o}f-\textbf{T}houghts \textbf{D}istillation (\textbf{AoTD}), a method that enhances models by incorporating automatically generated Chain-of-Thoughts (CoTs) into the instruction-tuning process. Specifically, we leverage an agent-based system to decompose complex questions into sub-tasks, and address them with specialized vision models, the intermediate results are then treated as reasoning chains. 
We also introduce a verification mechanism using a large language model (LLM) to ensure the reliability of generated CoTs. Extensive experiments demonstrate that AoTD improves the performance on multiple-choice and open-ended benchmarks.
\end{abstract}

\section{Introduction}
\label{sec:intro}
Video Question Answering (VideoQA) refers to a critical task that offers a natural interface for human-machine interaction through language~\cite{yu2019activitynet,wu2021star,xiao2021next,patraucean2023perception}. 
This synergy of visual content and language enhances the accessibility of AI systems for the general public, 
allowing users to query complex visual content with natural language. By encompassing tasks such as action recognition, object detection, and scene understanding, VideoQA serves as a comprehensive benchmark for evaluating AI’s ability to interpret videos, addressing the fundamental questions of `who', `what', `when', and `where' that are crucial to understand daily life activities, 
pushing the boundaries of what AI systems can interpret from dynamic visual content.

Recent literature has primarily explored two avenues in VideoQA. 
The first involves training large video language models (Video-LLMs) through direct instruction-tuning, using videos paired solely with corresponding questions and answers~\cite{alayrac2022flamingo,lin2023videollava,Maaz2023VideoChatGPT,cheng2024videollama}. While these models excel on public benchmarks, they often lack explainability and struggle with spatial-temporal grounding. This limitation hinders their ability to provide clear reasoning, which is essential for real-world applications where transparency and interpretability are critical~\cite{mitra2023orca}.

Conversely, an emerging approach utilizes agent-based systems that decompose complex questions into manageable sub-tasks, each addressed by specialized tools~\cite{suris2023vipergpt,gupta2023visualprogram,hu2024avis}.
The results are then aggregated to form a coherent answer. 
Theoretically, such approach naturally offers greater interpretability, 
as the reasoning process is divided into explainable steps that can be independently assessed. However, our experiments indicate that current video understanding tools are not strong enough for building reliable agent-based systems. In addition, the high memory demands and time-consuming nature of these systems present significant challenges for their practical use.

In this paper, we aim to leverage the advantage of both research lines,
enhancing Video-LLM by integrating Chain-of-Thoughts (CoTs) into instruction-tuning, with the CoTs being constructed from the outputs of specialized agent models, capturing the step-by-step reasoning procedure, as illustrated in Figure~\ref{fig: teaser}.

In specific, we start by systematically evaluating the off-the-shelf models tailored for atomic video understanding tasks, such as action recognition~\cite{weng2023open,wang2024internvideo2} and language grounding~\cite{lin2023univtg}, using well-annotated datasets. 
This comprehensive evaluation allows us to pinpoint the most effective tools for each sub-task, thus laying a robust foundation for constructing reliable chains.
Moreover, this process also provides a critical assessment of the broader capabilities of visual models across general and complex scenes, offering valuable insights for future research within the community.

In addition, we introduce a verification mechanism using a large language model (LLM), designed to assess if the generated CoTs adhere to a clear, step-by-step reasoning process and incorporate essential information for answering the queries effectively. This mechanism filters out low-quality or logically inconsistent reasoning paths. 
The remaining CoTs that pass this verification are then distilled into large generative video-language models, significantly enhancing both their performance and interpretability,
ultimately leading to the development of more robust, accurate, and interpretable VideoQA systems.

In summary, our contributions are three-fold:
(i) we propose a novel approach for enhancing Video-LLMs by distilling high-quality CoTs into their instruction-tuning process. These CoTs capture step-by-step reasoning paths, improving both the model’s performance and its interpretability;
(ii) to automatically construct the CoTs for any dataset, we employ an agent-based system to decompose complex VideoQA questions into simpler sub-tasks, leveraging off-the-shelf vision models to handle each sub-task. The intermediate outputs from these models can therefore be collected as CoTs for addressing the corresponding visual question; 
(iii) through extensive experiments, we demonstrate that our distilled model outperforms existing methods across both multiple-choice and open-ended VideoQA benchmarks, enabling to deliver not only accurate answers but also comprehensive reasoning explanations.

\begin{figure*}[t]
  \centering
  \resizebox{\linewidth}{!}{
  \includegraphics{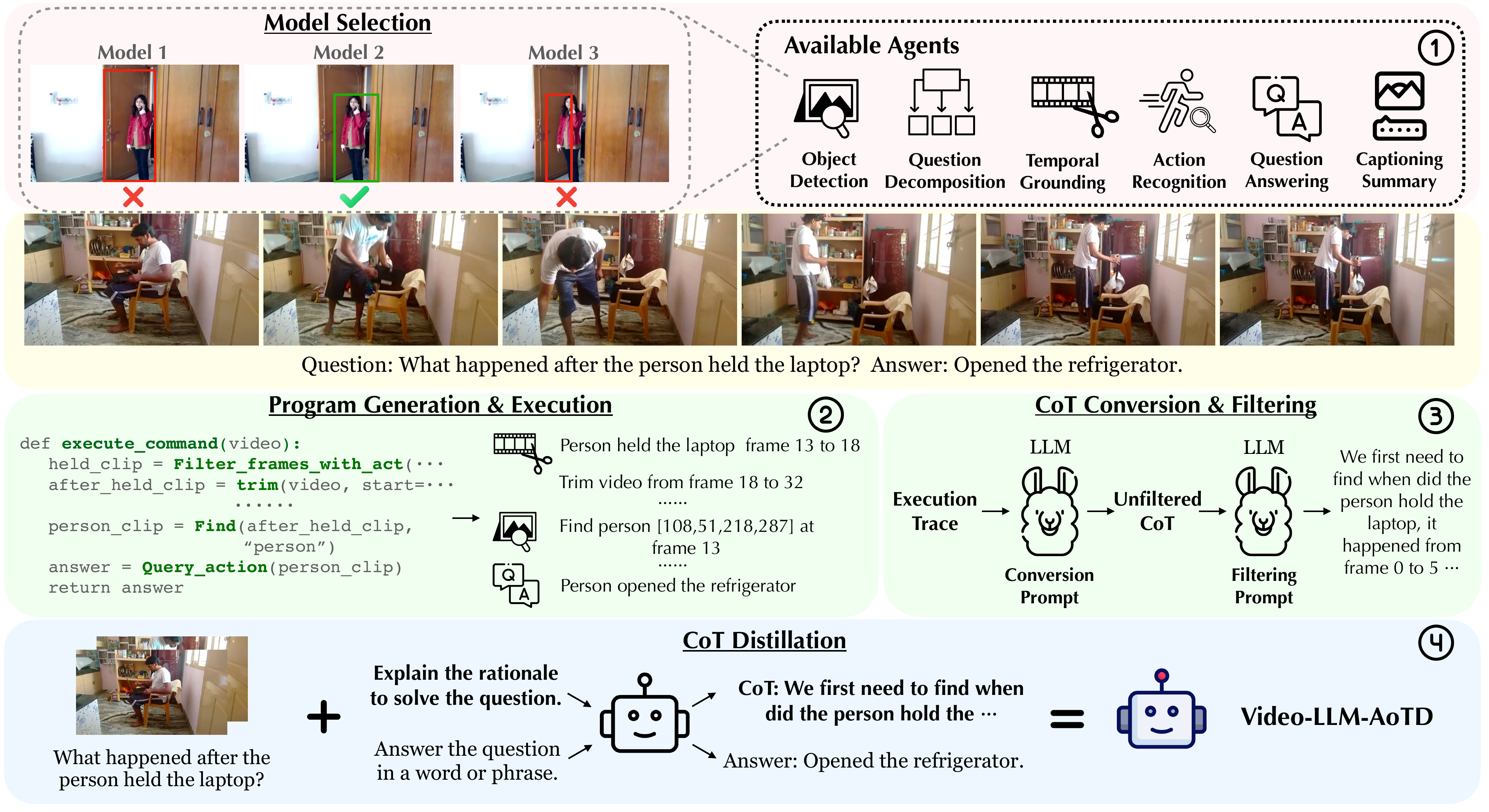}
  }
  \vspace{-15pt}
  \caption{
    Overview on Agent-of-Thoughts Distillation (AoTD). \textbf{Step 1: }Selecting best-performing agents for each sub-task to construct an agent-based system. \textbf{Step 2: }Decomposing question into executable program and leveraging chosen models to solve it sequentially to generate execution trace. \textbf{Step 3: }The execution trace is converted and filtered by LLM to produce high quality natural language CoTs. \textbf{Step 4: }Distilling CoTs into Video-LLM with two forms of prompt, allowing it achieve a balance between concise answers and comprehensive rationales. The final model is Video-LLM-AoTD.
  }  
  \label{fig: pipeline}
  \vspace{-1.1em}
\end{figure*}

\section{Related Work}
\noindent \textbf{Video-language models (Video-LLMs).} 
Most existing Video-LLMs are composed of a pre-trained visual encoder (like CLIP~\cite{radford2021clip} or SigLIP~\cite{zhai2023siglip}) to encode video frames into a sequence of visual features, an adapter to transfer the visual features to tokens that can be understood by the language model, and a pretrained LLM to output the final response. %These models achieve a strong ability for general vision-language tasks, such as video question-answering. 
Recent works such as VideoLLaMA2~\cite{damonlpsg2024videollama2}, LLaVA-NeXT-Video~\cite{zhang2024llavanextvideo} and VideoChat2~\cite{li2024mvbench}, with their excellent architecture design and reasonable instruction-tuning data collection, have achieved a new level of zero-shot results in VideoQA task. However, current end-to-end models still lack interpretability for questions, as well as the ability to think and visually process complex problems in multiple steps, %which leads to their weakness in real scenarios, 
which is an important part for embodied learning and autonomous driving.

\vspace{2pt}\noindent\textbf{Visual Programming and Agents.} With the progress of LLMs, some recent works~\cite{gupta2023visualprogram,suris2023vipergpt,choudhury2023proviq,yang2024doraemongpt} begin to try to use LLM as planner to solve the complex reasoning task in real scenarios. They attempt to decompose the question into some easier sub-questions, and use different specialist models as agents to solve these sub-questions, and finally gather them to get the answer of the raw question. MoReVQA \cite{min2024morevqa} proposes a multi-stage system, getting a strong zero-shot VideoQA ability while is able to create interpretable intermediate outputs. VURF \cite{mahmood2024vurf} proposes a self-refinement method to resolve the LLM hallucinations to get a more concise program based on the context cues. These models demonstrate a strong ability to obtain trustworthy answers based on the intermediate evidence they get, but they lag far behind the end-to-end model in terms of inference speed, and often require some in-context examples to assist them in solving problems, which brings a lot of trouble to the use of these agent-based models.

\vspace{2pt} \noindent\textbf{Visual Chain-of-Thoughts~(CoTs).}
The potential of Chain-of-Thought (CoT) reasoning~\cite{wei2022chain, yao2024tree} extends from NLP to the visual domain, highlighting a growing interest in applying this approach across various fields. Numerous studies have incorporated CoTs into visual understanding tasks~\cite{zhang2023mcot, mitra2024ccot, shao2024viscot, gao2024cantor}, utilizing powerful Multi-Modal Large Language Models (MLLMs) for generating CoTs or adopting tool-based architectures for sequential problem-solving. However, these methods encounter several limitations, such as errors in CoT generation by MLLMs and high time and memory costs for tool-based systems.

Recent innovations, for example, Visual Program Distillation~(VPD)~\cite{hu2024vpd} and Fact~\cite{gao2024fact} attempt to address these issues by maintaining the accuracy and diversity of CoTs, while leveraging MLLMs to generate them directly. These approaches decompose complex tasks into code programs, call upon expert models to handle sub-tasks, and utilize the resulting CoTs as training data to fine-tune visual-language models. This process significantly improves the models’ ability to generate detailed rationales.
Despite the progress in image understanding, there remains a notable oversight in video domains, where reasoning chains can be particularly effective due to the complex spatial-temporal dynamics of video understanding tasks. This is the focus of our paper.

\vspace{2pt} \noindent\textbf{Concurrent Work.}
In the recent literature, we notice two work that share similar idea with ours, specifically, Video-STaR~\cite{zohar2024videostar} construct CoTs using videos and existing labels for instruction-tuning, yet they do not develop an agent-based system. Meanwhile, MotionEpic~\cite{fei2024videothought} introduces a Video-of-Thought reasoning framework that integrates video spatial-temporal scene graphs, marking a significant stride towards more nuanced video reasoning.

\section{Agent-of-Thoughts Distillation} \label{sec: distillation}
In this paper, we propose a novel approach, termed Agent-of-Thoughts Distillation (AoTD), to enhance the Video-LLMs by training them with multi-step Chain-of-Thoughts. Specifically, we start by developing an agent-based video understanding system, 
to generate multi-step reasoning chains that address complex video questions. 
These reasoning chains are then distilled into one Video-LLM through instruction-tuning. By combining the strengths of agent-based systems and large generative models, our proposed AoTD enables to build  more reliable and interpretable VideoQA systems. Figure~\ref{fig: pipeline} illustrates the entire process of our method.

\begin{figure*}[t]
  \resizebox{\linewidth}{!}{
  \includegraphics{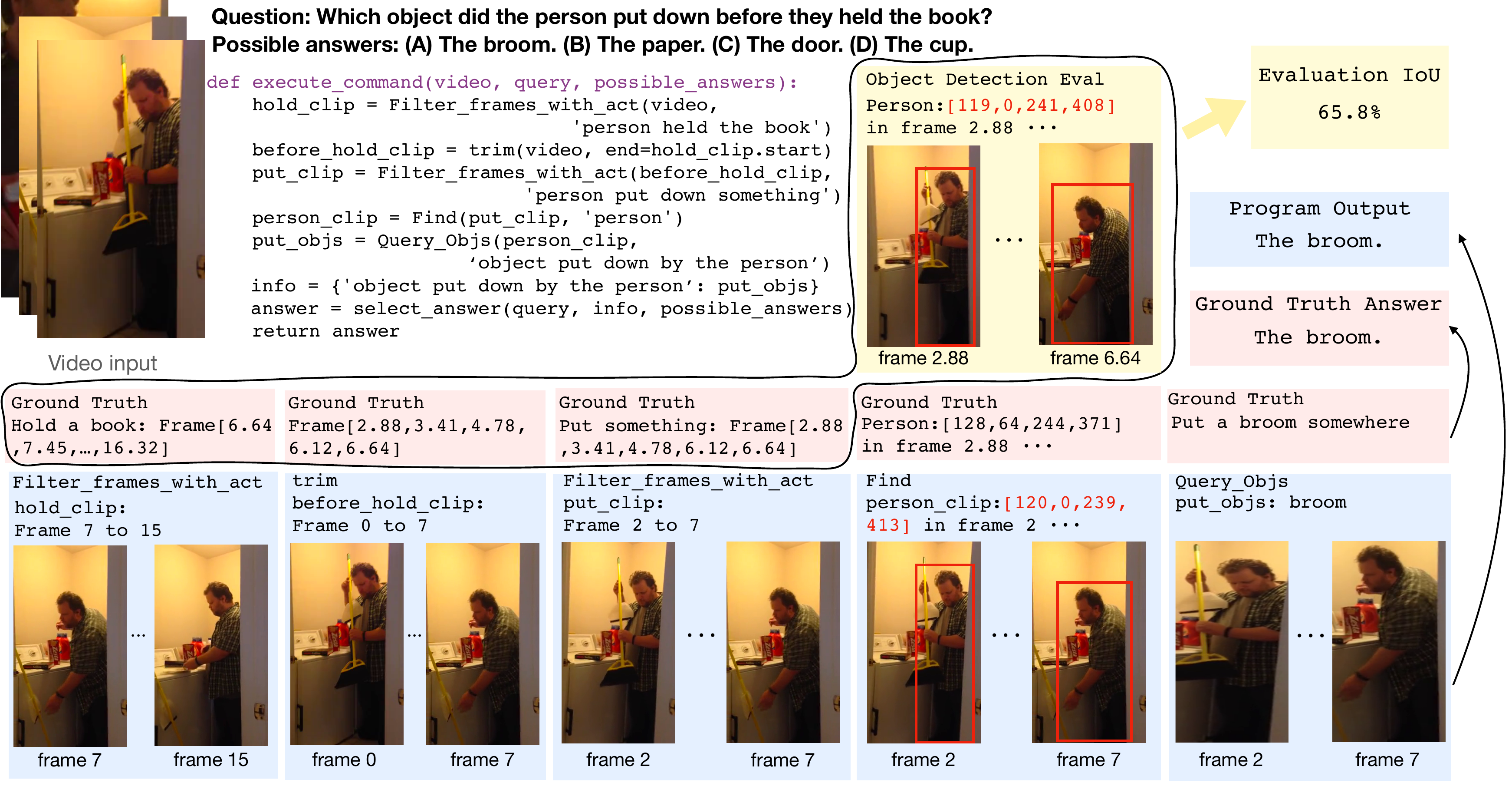}
  }
  \vspace{-15pt}
  \caption{
    Program execution process in an agent-based system. 
    We uniformly sample 32 frames from the video, and to ensure scale consistency, the frame ids of key frames are normalized into these 32 frames. 
    The blue boxes represent the program execution steps, 
    the red boxes denote the ground truth for each step.
    The combination of red and yellow boxes represents one example process of evaluating Object detection model candidates.
    }
  \label{fig: execute program}
  \vspace{-1.1em}
\end{figure*}

\subsection{Problem Formulation}

Given a video clip with $t$ frames, $\mathcal{V} = \{x_1, \dots, x_t\}$, 
and a set of $n$ questions $\mathcal{Q} =\{q_1, q_2, . . . , q_n \}$, 
our goal is to train a Video-LLM capable of producing both concise answers and comprehensive rationales. Depending on the suffix prompt $p_s$, the model can generate different types of outputs. The process can be formulated as:
\begin{align*}
    \{a_i, \mathcal{S}_i\} = \Phi(\mathcal{V}, q_i, p_s), \text{\hspace{3pt}}
    \mathcal{S}_i = \{\emptyset\} \text{ or } \{s_{i,1}, \dots, s_{i, k}\}
\end{align*}
where $q_i$ denotes the $i$-th question, $a_i$ is the answer in free-form text, 
and $\mathcal{S}_i$ represents the rationale, consisting of the reasoning process. If the prompt specifies to only generate the answer, $\mathcal{S}_i = \{\emptyset\}$. Otherwise, if the prompt requires the generation of rationales, $\mathcal{S}_i = \{s_{i,1}, \dots, s_{i,k}\}$, where each $s_{i,j}$ corresponds to a reasoning step.

\vspace{3pt} \noindent \textbf{Discussion.} 
Unlike existing models that are instruction-tuned on VideoQA datasets using simple question-answer pairs, which bypass the intermediate thought process, our approach emphasizes the importance of training with Chain-of-Thoughts (CoTs). 
In the following section, we outline the process for generating high-quality CoTs from existing VideoQA datasets.

\subsection{CoT Construction with Agent-based System}
\label{sec: agent-based system}
Recent work, such as STAR \cite{wu2021star}, has introduced the idea of employing executable symbolic programs, to directly decompose questions into sub-tasks. 
When combined with scene graphs that contain comprehensive video information from key frames—such as object locations, interactions, and actions—these programs facilitate the generation of concise Chain-of-Thoughts (CoTs) through direct execution of symbolic operations. However, datasets of this nature are limited in scale, 
we therefore propose to first build an agent-based system, 
capable of breaking down complex questions into simpler sub-tasks, 
and the intermediate outputs from this system can then be employed to construct CoTs for any existing VideoQA dataset.

\begin{table}[t]
  \centering
  \footnotesize
  % \scriptsize
  % \small
  \renewcommand{\arraystretch}{1.2} % Adjust row height
  \setlength{\tabcolsep}{0.1mm}      % 控制列间距
  % \vspace{1em}
  \resizebox{\linewidth}{!}{
  \begin{tabular}{clcc}
    \toprule
    Sub-task & Model name & Metric &  Number (\%)\\
    \midrule
    \multirow{3}{*}{\parbox{2cm}{\centering question \\ decomposition}} 
                           & CodeQwen1.5-Chat (7B) \cite{qwen} &    & 52.7 \\
                           & GPT-3.5-Turbo \cite{openai2023gpt3.5turbo} & Acc & 73.1 \\
                           & \cellcolor{gray!20} DeepSeek-Coder-Instruct (6.7B) \cite{deepseek-coder} &\cellcolor{gray!20}   & \cellcolor{gray!20} 85.7 \\
    \hline
    \rule{0pt}{10pt}
    \multirow{3}{*}{\parbox{2cm}{\centering object \\ detection}} 
                           & OWL-ViT v1 \cite{minderer2022simple} &    & 47.3\\
                           & GLIP \cite{li2021glip}     & IoU & 58.9\\ % Centering IoU
                           & \cellcolor{gray!20} OWL-ViT v2 \cite{minderer2024scaling}  &  \cellcolor{gray!20}  & \cellcolor{gray!20} 63.0\\
    \hline
    \rule{0pt}{10pt}
    \multirow{3}{*}{\parbox{2cm}{\centering temporal \\ grounding}} 
                           & LITA (13B) \cite{huang2024lita} &    & 11.7 / 20.2 \\
                           & TimeChat (7B) \cite{ren2024timechat} & IoU / Recall & 13.9 / 23.1 \\ % Centering IoU
                           % & BLIP2 &    & 23.8 / 98.5 \\
                           & \cellcolor{gray!20} UniVTG \cite{lin2023univtg} & \cellcolor{gray!20}   & \cellcolor{gray!20} 24.7 / 35.3 \\
    \hline
    \rule{0pt}{10pt}
    \multirow{3}{*}{\parbox{2cm}{\centering action \\ recognition}} 
                           & InternVideo2 (1B) \cite{wang2024internvideo2} &  & 7.6  \\%\multirow{2}{*}{\centering Top1-Acc}  & 8.9 \\
                           & Open-VCLIP \cite{weng2023open} & Top1-Acc & 8.9 \\
                           & \cellcolor{gray!20} LLaVA-NeXT-Video-DPO (7B) \cite{zhang2024llavanextvideo}        & \cellcolor{gray!20} & \cellcolor{gray!20} 18.2    \\
    \hline
    \rule{0pt}{10pt}
    \multirow{3}{*}{\parbox{2cm}{\centering question \\ answering}} 
                           & LLaMA-VID (7B) \cite{li2024llamavid} &    & 43.5 \\
                           & SeViLA \cite{yu2023sevila} & Acc & 46.5 \\
                           & \cellcolor{gray!20} LLaVA-NeXT-Video-DPO (7B) \cite{zhang2024llavanextvideo} &  \cellcolor{gray!20}  & \cellcolor{gray!20} 53.4 \\
    \bottomrule
  \end{tabular}
  }
  \vspace{-5pt}
  \caption{
    Sub-tasks definition and evaluation results. 
    We choose 3 model candidates for each sub-task and evaluate them in STAR training set with the corresponding metrics. 
    Models with best performance are placed at the bottom of each column.
  }
  \vspace{-1.5em}
  \label{tab: sub-tasks evaluation results.}
\end{table}

\vspace{3pt} \noindent \textbf{Agent-based VideoQA.}
Assuming we are given a video input~($\mathcal{V}$), questions~($\mathcal{Q}$), 
and a set of visual models~($\mathcal{M} = \{ \phi_{\text{act}}, \phi_{\text{det}}, \ldots, \phi_{\text{qa}}\}$), an LLM-based agent core~($\pi(\cdot)$) processes the question along with the documentation of the visual models~($\mathcal{T}$), which includes variables and functionalities. 
The agent subsequently decomposes the question into sub-tasks formatted as Python code, and resolves them by invoking the appropriate visual models through function calls. It is important to note that the visual models can be arranged in various orders depending on the specific question, ensuring flexibility in problem-solving.

Specifically, as illustrated by the example in Figure~\ref{fig: execute program}, 
the question is first decomposed into a series of sub-tasks, including temporal grounding, object detection, and question answering. The corresponding specialized models are then executed sequentially to address these sub-tasks, ultimately yielding the final answer $y_i$:
\begin{align*}
&\{\phi_{\text{ground}}, \phi_{\text{det}}, \phi_{\text{qa}}\} := \pi(q_i, \mathcal{T}), \\
&y_i = \phi_{\text{ground}}(\mathcal{V}) \rightarrow \phi_{\text{det}}(\mathcal{V}) \rightarrow \phi_{\text{qa}}(\mathcal{V})
\end{align*}

\noindent \textbf{CoT Construction.}
To ensure the correctness of outputs at all the intermediate steps, 
we leverage the training set from STAR for hyperparameter tuning, 
enabling us to identify the most effective model for each sub-task within the agent-based system. By following the provided programs, we evaluate the performance of the corresponding vision models on tasks such as object detection and action recognition. Given the availability of complete reasoning chains, we independently assess each sub-task using ground truth data for all preceding steps.

As shown in Table~\ref{tab: sub-tasks evaluation results.}, we present the evaluation results for the various sub-tasks. Specifically, for \textbf{question decomposition}, we compare several code LLMs, with DeepSeek-Coder-Instruct achieving the highest performance, outperforming even GPT-3.5-Turbo. In \textbf{object detection}, OWL-ViT v2 records the highest Intersection over Union (IoU) score, showcasing its superior open-vocabulary detection capability. The results for \textbf{temporal grounding} indicate that while UniVTG leads in performance, there remains a need for further advancements in this area. In \textbf{action recognition}, our evaluations show that generative models outperformed discriminative models, likely due to the fine-grained action list provided by the STAR dataset. However, the performance of both model types reveals significant room for improvement. Finally, in the \textbf{one-hop question answering} sub-task, all models perform admirably, with LLaVA-NeXT-Video-DPO standing out as a top performer, consistent with its strong results on other benchmarks.

With these high-performing models, we implement the agent-based approach on VideoQA datasets that consist solely of QA pairs. During the execution of the programs, we record all intermediate outputs to construct the CoTs. Since the outputs from these vision models vary in format—such as bounding boxes and free-form text—we employ another LLM to translate the execution trace into natural language for better use in the distillation process. Detailed examples are provided in Appendix~\ref{sec: cot visualization}.

\subsection{CoT Verification}
To refine the quality of reasoning chains for VideoQA samples, 
we implement a two-step verification: 
(i) we filter execution traces to retain only those, 
where the program can reach correct output. 
For multiple-choice datasets, the output must match the correct answer exactly, 
while for open-ended datasets, we prompt the LLM to verify correctness, accounting for format differences; 
(ii) we prompt the LLM to evaluate the logical coherence and usefulness of the reasoning chains in solving the problem. The model assesses whether the CoTs follow a clear, step-by-step reasoning process and provides a binary evaluation (`Yes' or `No') to indicate their quality (detailed prompts are included in Appendix~\ref{sec: prompts}). This two-step approach ensures that only high-quality CoTs are retained for further distillation.

In Table \ref{tab: cot static.}, we provide the statistics for the remaining generated CoTs for different datasets. We primarily select compositional QA datasets, as these require the model to process spatial-temporal information from different events comprehensively. 

\begin{table}[t]
\centering
    \footnotesize
    \renewcommand{\arraystretch}{1.2} % Adjust row spacing here
    \begin{tabular}{llrr}
    \toprule
    \textbf{Dataset} & \textbf{Description} & \textbf{\# Labels} & \textbf{\# CoTs} \\ \hline
    AGQA & Compositional & 25.0K   & 5.4K \\ 
    ANetQA & Compositional  & 25.0K  & 3.6K  \\ 
    STAR & Compositional & 45.7K  & 11.2K  \\ 
    NExT-QA & Temporal \& Causal  & 34.1K  & 12.1K   \\ 
    CLEVRER & Spatial \& Temporal    & 21.0K     & -         \\ 
    EgoQA & Ego-centric  & 7.8K  & -    \\ 
    \hline
    \textbf{Total}  &  & \textbf{158.6K}  & \textbf{32.3K}  \\
    \bottomrule
    \end{tabular}
    \vspace{-5pt}
    \caption{Dataset statistics. The column ``\# Labels" indicates the number of VideoQA pairs, which include the video, query, possible answers (multiple-choice), and the correct answer. ``\# CoTs" refers to the number of CoTs generated using our agent-based system for each dataset.}
    \label{tab: cot static.}
    \vspace{-1em}
\end{table}

\begin{figure*}[t]
  \resizebox{\linewidth}{!}{
  \includegraphics{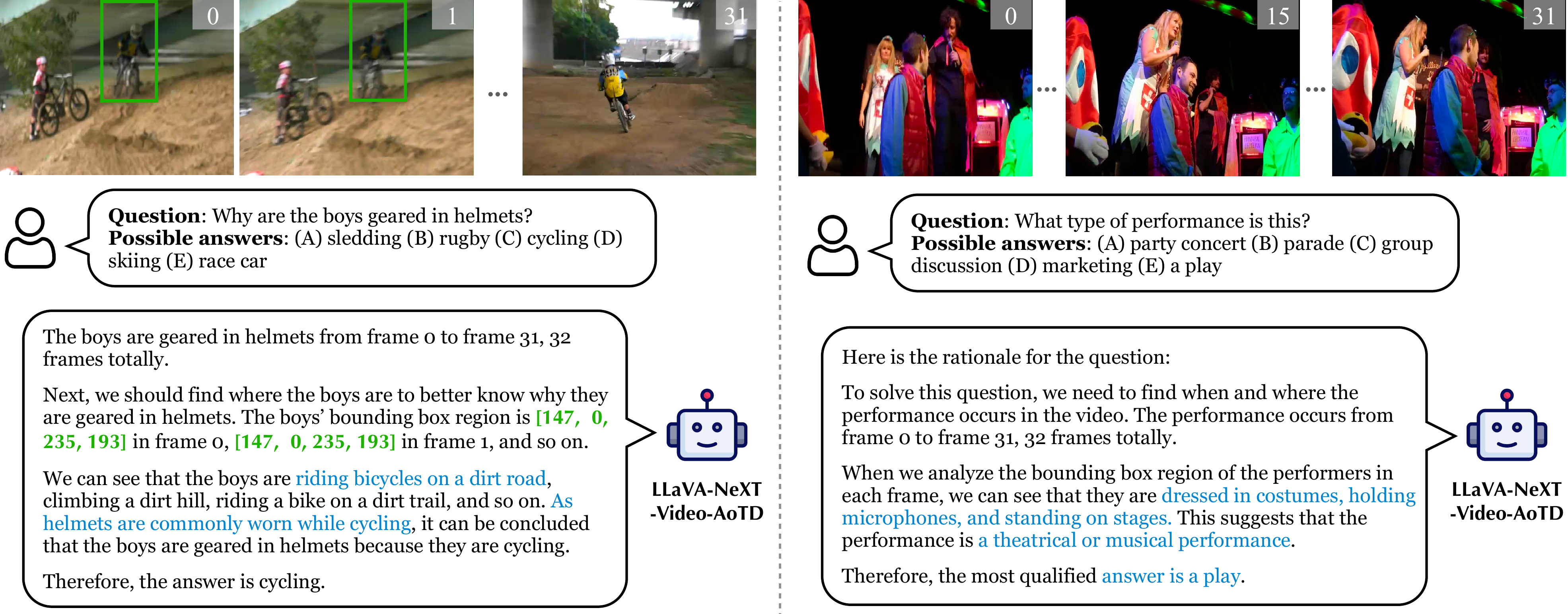}
  }
  \vspace{-15pt}
  \caption{
    \textbf{Visualization of rationales.} LLaVA-NeXT-Video-AoTD can output rationales containing both spatial-temporal grounding of key information and step-by-step thinking process to solve the question.
    }
  \label{fig: visualization}
  \vspace{-1em}
\end{figure*}

\subsection{Step-by-step Distillation}
In this section, we describe the process of distilling the generated CoTs into a Video-LLM. This distillation enhances the model's ability for spatial-temporal video understanding and multi-step reasoning, thereby improving its performance on complex VideoQA tasks.

In specific, using the generated CoTs, we can build the dataset $D = \{(\mathcal{V}_j,q_j,\hat{y}_j,c_j,p_s)\}_{j=1}^N$, 
where $N$ is the total number of samples in the distilling dataset, $\mathcal{V}_j$ is the video input, $q_j$ is the question, $\hat{y}_j$ is the ground-truth answer, $c_j$ is the generated CoT, $p_s$ is the task-specific suffix prompt, to distinguish different tasks, for example, for multiple-choice VQA, the prompt can be:
``Answer with the option's letter from the given choices directly and only give the best option", and for open-ended VQA, the prompt can be: ``Answer in one word or phrase". 
Please refer to detailed prompts in Appendix~\ref{sec: prompts}.

At distillation stage, we minimize the cross-entropy loss of predicting both the answer and the CoTs, we replace the suffix prompt $p_s$ with ``Explain the rationale to answer the question", to indicate whether we want a question answer or a rationale to explain the thinking steps. 
Following \cite{gao2024fact} and \cite{hu2024vpd}, our optimization objective is:
\begin{align*}
\mathcal{L}&=\mathcal{L}_{\text{label}}+\lambda\mathcal{L}_{\text{rationale}} \\
&=\sum_{j=1}^N\ell(\Phi(\mathcal{V}_j,q_j,p_s),\hat{y}_j)+\lambda\ell(\Phi(\mathcal{V}_j,q_j,p_s),c_j)
\end{align*}
Here, we set $\lambda$ to 1 to ensure the importance of answer and rationale are equally considered. Notice that, not all the QA pairs can generate qualified CoT. In that case, the $\mathcal{L}_{\text{rationale}}$ will be set to 0.

\begin{table}[h]
\centering
\footnotesize
% \scriptsize
\renewcommand{\arraystretch}{1.1}
\setlength{\tabcolsep}{0.9mm}      % 控制列间距
% \vspace{-10em}
% \resizebox{\linewidth}{!}{
\begin{tabular}{lcclcc}
\toprule
\multirow{2}{*}{\textbf{Dataset}}& \multicolumn{2}{c}{\textbf{Size}} & \multirow{2}{*}{\textbf{Type}} & \multirow{2}{*}{\textbf{Train}} & \multirow{2}{*}{\textbf{Eval}} \\ 
& \textbf{train} & \textbf{eval} & & & \\
\hline
\multicolumn{6}{l}{\textbf{MC-VQA}} \\
STAR~\cite{wu2021star}& 45.7K & 7.1K & Compositional & \cmark & \cmark \\ 
NExT-QA~\cite{xiao2021next}& 34.1K & 5.0K & Temporal \& Causal & \cmark & \cmark \\ 
CLEVRER~\cite{CLEVRER2020ICLR}& 21.0K & - & Spatial-temporal & \cmark & \xmark \\ 
Perception-Test~\cite{patraucean2023perception}& - & 11.5K & General & \xmark & \cmark \\ 
MVBench~\cite{li2024mvbench}& - & 4.0K & General & \xmark & \cmark \\ 
VideoMME~\cite{fu2024videomme}& - & 2.7K & General & \xmark & \cmark \\ 
VSIBench~\cite{yang2024think}& - & 5.0K & spatial-temporal & \xmark & \cmark\\
\hline
\multicolumn{6}{l}{\textbf{OE-VQA}} \\
AGQA~\cite{grunde2021agqa}& 25.0K & 2.0K & Compositional & \cmark & \cmark\\
ANetQA~\cite{yu2023anetqa}& 25.0K & 2.0K & Compositional & \cmark & \cmark \\ 
EgoQA~\cite{grauman2022ego4d}& 7.8K & - & Ego-centric & \cmark & \xmark \\ 
Activitynet-QA~\cite{yu2019activitynet}& - & 8.0K & General & \xmark & \cmark \\ 
Video-ChatGPT~\cite{Maaz2023VideoChatGPT}& - & 3.0K & General & \xmark & \cmark \\ 
% \midrule
% \multicolumn{6}{l}{\textbf{Hybrid}} \\
\bottomrule
\end{tabular}
% }
\vspace{-5pt}
\caption{
      Training and evaluation datasets statics. 
      % \weidi{more details.}
    }
\label{tab: dataset static}
\vspace{-1.5em}
\end{table}

\begin{table*}[t]
\centering
\renewcommand{\arraystretch}{1.2} % Adjust row height
\setlength{\tabcolsep}{3.1mm}      % 控制列间距
\footnotesize
% \small
% \vspace{1em}
% \resizebox{\linewidth}{!}{
\begin{tabular}{lcccccc}
\toprule
\multirow{2}{*}{\textbf{Model}} & \textbf{MVBench} & \textbf{VideoMME} & \textbf{STAR} & \textbf{NExT-QA}& \textbf{Perception-Test} & \textbf{VSIBench}\\
 & \textbf{(Acc.)} & \textbf{(Acc.)} & \textbf{(Acc.)} & \textbf{(Acc.)} & \textbf{(Acc.)} & \textbf{(Acc.)}\\
\midrule
\multicolumn{7}{l}{\textbf{Proprietary Models}} \\
% Gemini 1.0 Pro \cite{team2023gemini} & - & - & - & - & 51.1\\
% Gemini 1.0 Ultra \cite{team2023gemini} & - & - & - & - & 54.7\\
% Gemini 1.5 Flash \cite{team2024gemini} & - & - & 70.3 & - & -\\
Gemini 1.5 Pro \cite{team2024gemini} & - & 75.0 & - & - & - & 45.4\\
GPT4-V \cite{openai2023gpt4v} & 43.7 & 59.9 & - & - & - & -\\
% GPT4-O \cite{openai2024gpt4o} & - & 71.9 & - & - & -\\
\midrule
\multicolumn{7}{l}{\textbf{Open-source Models}} \\
LLaMA-VID~(7B) \cite{li2024llamavid} & 41.9 & 25.9 & - & - & 44.6 & -\\
Video-LLaVA~(7B) \cite{lin2023videollava} & 41.0 & 39.9 & - & - & 44.3 & -\\
VideoChat2~(7B) \cite{li2024mvbench} & 51.1 & 33.7 & 59.0\rlap{\textsuperscript{*}} & 68.6\rlap{\textsuperscript{*}} & 47.3 & -\\
VideoLLaMA2~(7B) \cite{damonlpsg2024videollama2} & 53.4 & 45.1 & 58.5\rlap{\textsuperscript{*}} & 
62.3\rlap{\textsuperscript{*}} & 49.6 & -\\
LLaVA-NeXT-Video~(7B) \cite{zhang2024llavanextvideo} & 46.5\rlap{\textsuperscript{*}} & 41.0\rlap{\textsuperscript{*}} & 52.4\rlap{\textsuperscript{*}} & 61.6\rlap{\textsuperscript{*}} & 47.5\rlap{\textsuperscript{*}} & 19.7\rlap{\textsuperscript{*}}\\
% LLaVA-NeXT-Interleave (7B) & - & 53.1 & - & - & 78.2\\
\hline
\rowcolor{green!10} LLaVA-NeXT-Video-Instruct~(7B) & 53.4 & 43.2 & 72.2 & 77.1 & 50.3 & 26.7\\
\rowcolor{green!10} LLaVA-NeXT-Video-AoTD~(7B) & 55.6 & 45.0 & 74.3 & 77.6 & 50.6 & 28.8\\
\bottomrule
\end{tabular}
% }
\vspace{-5pt}
\caption{Comparison with Video-LLMs on MC-VQA benchmarks. LLaVA-NeXT-Video-AoTD outperforms all other baselines the and the version without CoT distillation. * means results reproduced by ourseleves. Results without signs are retrieved from \cite{damonlpsg2024videollama2} and \cite{yang2024think}.
}
\label{tab:mc performance}
\vspace{-1em}
\end{table*}

\section{Experiments} \label{sec: experiments}
In this section, we present the experimental setup~(Sec.~\ref{sec: experimental setup}) and results on various VideoQA benchmarks~(Sec.~\ref{sec: quantitative results}). Extensive ablation studies have also been conducted to further examine the contributions of our approach in Sec.~\ref{sec: ablation study}, and an evaluation on the quality of rationales generated by the distilled model is made in Sec.~\ref{sec: rationale evaluation}.

\subsection{Experimental Setup} \label{sec: experimental setup}
\noindent \textbf{Base model.}
We use LLaVA-NeXT-Video~(7B) \cite{zhang2024llavanextvideo} (LNV for short)
as base Video-LLM, which has shown remarkable performance on image-centric tasks, for example image question answering~\cite{yue2024mmmu}.
We present comparison on naive instruction-tuning with video question answering dataset or with additional CoT distillation. For CoT conversion and verification, we prompt LLaMA-3.1-8B with the manually-designed instruction and some in-context examples. Detailed prompts are provided in Appendix~\ref{sec: prompts}.

\vspace{3pt}\noindent \textbf{Instruction tuning.}
We utilize both multiple-choice and open-ended QA data, along with the generated CoTs, to fine-tune the base video question answering model, as summarised in Table~\ref{tab: cot static.}. The resulting distilled model is named \textbf{LLaVA-NeXT-Video-AoTD} (LNV-AoTD for short). Additionally, as baseline, we also train another version of the model using only the basic QA data, which we refer to as \textbf{LLaVA-NeXT-Video-Instruct} (LNV-Instruct for short).

\vspace{3pt}\noindent \textbf{Evaluation benchmarks.}
We conduct extensive evaluations on Multiple-Choice Video QA (MC-VQA) and Open-Ended Video QA (OE-VQA). We report the top-1 accuracy for all MC benchmarks, which means the proportion of the output equal to the answer. 
We report a GPT-assessed Acc. and Score with the help of GPT-3.5-turbo-0613 for all OE benchmarks. 
For each question, GPT delivers a binary decision indicating whether the output is correct or incorrect, along with a similarity score reflecting the degree of alignment between the output and the correct answer. The term `Acc.' refers to the percentage of correct outputs, while `Score' represents the average similarity scores.
For the evaluation on AGQA and ANetQA, due to the large volume of test set, we test on a subset of samples. We evenly select the benchmark in-domain and out-of-domain for testing to ensure a comprehensive and reasonable evaluation of the model capability. Noted that though VSIBench has both MC and OE questions, it doesn't need GPT for score, so we classify it into MC benchmarks for convenience. Detailed statistics for evaluation benchmarks are shown in Table~\ref{tab: dataset static}. 

\subsection{Quantitative Results}\label{sec: quantitative results}
We divide the comparison into two parts: the first focuses on comparing the distilled model with other baselines, while the second examines the difference between the instruct version and the AoTD version. 
Note that, the latter part will be mainly compared and discussed, 
to demonstrate the model's improvement relative to its previous performance, as well as establishing the transferability of the method across models.

\vspace{3pt} \noindent \textbf{MC-VQA performance.}
As shown in Table~\ref{tab:mc performance}, 
our LLaVA-NeXT-Video-AoTD achieves superior performance across all benchmarks. Several key observations can be made: 
(i) comparing to the base model, even a simple instruction-tuning on certain VideoQA datasets significantly enhances the model's question-answering performance. This improvement is notable, as the base model was primarily trained on static images and struggled with video understanding;
(ii) our model, instruction-tuned with CoT distillation, demonstrates further performance enhancements across all benchmarks, particularly on the compositional VideoQA benchmark (STAR) and comprehensive benchmarks (VideoMME, MVBench). This suggests that our AoTD method effectively improves the model's ability to address complex problems and interpret spatial-temporal scenes; 
(iii) the distilled model consistently outperforms all other baselines across almost all benchmarks, even when compared to more powerful models. This finding shows that our method effectively bridges 
performance gaps created by varying model components.

\begin{table*}[t]
    \centering
    \renewcommand{\arraystretch}{1.2} % Adjust row height
    \setlength{\tabcolsep}{1.7mm}      % 控制列间距
    \footnotesize
    % \resizebox{\linewidth}{!}{
    \begin{tabular}{lcccccccc}
    \toprule
         \multirow{2}{*}{\textbf{Model}} & \textbf{ANetQA} & \textbf{AGQA} &  \multicolumn{5}{c}{\textbf{Video-ChatGPT (Score)}}& \textbf{ActivityNet}\\
        & \textbf{(Acc./Score)} & \textbf{(Acc./Score)} & \textbf{Corr.} & \textbf{Deta.} & \textbf{Cont.} & \textbf{Temp.} & \textbf{Cons.} & \textbf{(Acc./Score)} \\
        \midrule
        \multicolumn{9}{l}{\textbf{Proprietary Models}} \\
        % Gemini 1.0 Pro \cite{team2023gemini} & - & - & - & - & - & - & - & 49.8/- \\
        % Gemini 1.0 Ultra \cite{team2023gemini} & - & - & - & - & - & - & - & 52.2/- \\
        Gemini 1.5 Pro \cite{team2024gemini} & - & - & - & - & - & - & - & 56.7/- \\
        GPT4-V \cite{openai2023gpt4v} & - & - & 4.09 & 3.88 & 4.37 & 3.94 & 4.02 & 59.5/- \\
        % GPT4-O \cite{openai2024gpt4o} & - & - & - & - & - & - & - & 61.9/- \\
        \midrule
        \multicolumn{9}{l}{\textbf{Open-Source Models}} \\
        VideoLLaMA~(7B) \cite{cheng2024videollama} & - & - & 1.96 & 2.18 & 2.16 & 1.82 & 1.79 & 12.4/1.1 \\
        Video-ChatGPT~(7B) \cite{Maaz2023VideoChatGPT} & - & - & 2.50 & 2.57 & 2.69 & 2.16 & 2.20 & 35.2/2.7 \\
        LLaMA-VID~(7B) \cite{li2024llamavid} & - & - & 2.96 & 3.00 & 3.53 & 2.46 & 2.51 & 47.4/3.3 \\
        Video-LLaVA~(7B) \cite{lin2023videollava} & - & - & 2.87 & 2.94 & 3.44 & 2.45 & 2.51 & 45.3/3.3 \\
        VideoChat2~(7B) \cite{li2024mvbench} & - & - & 3.02 & 2.88 & 3.51 & 2.66 & 2.81 & 49.1/3.3 \\
        VideoLLaMA2~(7B) \cite{damonlpsg2024videollama2} & - & - & 3.09 & 3.09 & 3.68 & 2.63 & 3.25 & 49.9/3.3 \\
        LLaVA-NeXT-Video~(7B) \cite{zhang2024llavanextvideo} & 46.4/3.3\rlap{\textsuperscript{*}} & 27.4/2.2\rlap{\textsuperscript{*}} & 3.26\rlap{\textsuperscript{*}} & 3.22\rlap{\textsuperscript{*}} & 3.77\rlap{\textsuperscript{*}} & 2.47\rlap{\textsuperscript{*}} & 2.99\rlap{\textsuperscript{*}} & 54.3/3.2\rlap{\textsuperscript{*}}\\
        % LLaVA-NeXT-Interleave (7B) & 55.3/3.1 & - & - & 3.51 & 3.28 & 3.89 & 2.77 & 3.68 \\
        \hline
        \rowcolor{green!10}
        LLaVA-NeXT-Video-Instruct~(7B) & 47.1/3.1 & 59.3/3.4 & 2.96 & 2.81 & 3.35 & 2.42 & 2.82 & 50.0/3.3\\
        \rowcolor{green!10}
        LLaVA-NeXT-Video-AoTD~(7B) & 53.9/3.4 & 60.9/3.6 & 3.11 & 3.00 & 3.60 & 2.41 & 2.91 & 53.2/3.4\\
        \bottomrule
    \end{tabular}
    % }
    \vspace{-5pt}
    \caption{Comparison with Video-LLMs on OE-VQA benchmarks. LLaVA-NeXT-Video-AoTD improves performance in all open-ended benchmarks compared with the Instruct version. * means results reproduced by ourseleves. Results without signs are retrieved from \cite{damonlpsg2024videollama2}.}
    \label{tab:oe performance}
    \vspace{-1em}
\end{table*}

\vspace{3pt} \noindent \textbf{OE-VQA performance.}
As shown in Table \ref{tab:oe performance}, LLaVA-NeXT-Video-AoTD outperforms the Instruct variant across all open-ended VideoQA benchmarks. 
Notably, it achieves a greater percentage increase compared to the MC-VQA benchmarks, suggesting that CoT distillation may be more effective for open-ended generation than for multiple-choice selection.
While the distilled model scores higher than most models listed in the table, it does not surpass LLaVA-NeXT-Video on certain benchmarks. 
We conjecture this is due to the model's extensive training on images, 
that can also benefit the question answering without requiring  complex reasonings, as also suggested by the findings in VideoLLaMA2~\cite{damonlpsg2024videollama2}. 
Additionally, the inherent challenges of evaluating open-ended VQA may influence the results. Assessments conducted by GPT can be biased or inaccurate, and the metrics we employ primarily indicate general trends rather than providing absolute accuracy.

\subsection{Ablation Study}\label{sec: ablation study}
\noindent \textbf{Analysis on CoT filtering.}
To prove the effectiveness of our filtering mechanism, we trained an alternative model without CoT filtering while maintaining all other settings,
{\em i.e.}, using 36.3K verified CoTs for distillation.
As shown in Table~\ref{tab: ablation study}, the model's performance declines significantly on both the Multiple-Choice VQA and Open-Ended VQA benchmarks when the CoT filtering mechanism is not utilized. This confirms that employing large language models (LLMs) to filter CoTs is crucial for enhancing data quality.

\vspace{3pt} \noindent \textbf{Analysis on model transferability.}
As AoTD is a distillation method that leverages Chain-of-Thoughts (CoTs), 
it can theoretically be applied to any Video-LLMs. To assess the transferability of our method, we conduct experiments on another very recent model, LLaVA-OneVision (7B) \cite{li2024onevision}. As shown in Table~\ref{tab: ablation study}, 
our method also demonstrates significant improvements on the benchmarks, 
showing the transferability and robustness of the approach. 
Due to the rapid advancements in the computer vision field, 
evaluating all models and benchmarks is prohibitively infeasible. 
Thus, we focus on assessing some representative models against selected benchmarks to provide a representative evaluation.

\vspace{3pt} \noindent \textbf{Analysis on latency and computation.}
We compare the agent-based system with the distilled model (LNV-AoTD) across three key metrics. As shown in Table~\ref{tab: latency and computation}, LNV-AoTD outperforms the agent-based system in both inference latency and computational efficiency, demonstrating the necessity of distillation process.

\begin{table}[ht]
    \vspace{-8pt}
    \centering
    \footnotesize
    % \small
    % \tiny
    % \scriptsize
    % \vspace{1em}
    % \setlength{\tabcolsep}{4pt}
    % \resizebox{0.95\linewidth}{!}{
    \begin{tabular}{lccc}
            \toprule
            {\textbf{Model}} & \textbf{Time~(s)} $\downarrow$& \textbf{Memory~(GB)} $\downarrow$& \textbf{TFLOPs} $\downarrow$\\
            %& \textbf{(s)} &
            %\textbf{(GB)} &
            %\textbf{(TFLOPs)} \\
            \midrule
            Agent-based system & 47.93 & 65.34 & 30.98\\
            LNV-AoTD & 10.58 & 18.21 & 13.53\\
            \bottomrule
        \end{tabular}
        \vspace{-5pt}
        \caption{Latency and computation comparison.}
        \label{tab: latency and computation}
        \vspace{-2em}
\end{table}

\begin{table}[ht]
    \centering
    \renewcommand{\arraystretch}{1.1} % 调整行距
    \footnotesize
    % \small
    % \vspace{1em}
    \setlength{\tabcolsep}{3pt}
    \begin{tabular}{lcccc}
            \toprule
            \multirow{2}{*}{\textbf{Model}} & \multirow{2}{*}{\textbf{Filtering}} & \textbf{MVBench} & \textbf{STAR} & \textbf{AGQA}\\
             & & \textbf{(Acc.)} & \textbf{(Acc.)} & \textbf{(Acc. / Score)}\\
            \midrule
            LNV-AoTD & \xmark & 53.7 & 73.3 & 59.5/3.5 \\
            LNV-AoTD & \cmark & 55.6 & 74.3 & 60.9/3.6 \\ \hline
            % Onevision & - & 58.0 & 65.9 & 39.0/3.0 \\
            Onevision-Instruct & - & 59.2 & 75.8 & 65.6/3.7 \\
            Onevision-AoTD & \cmark & 60.5 & 76.6 & 65.7/3.7 \\
            VideoLLaMA2-Instruct & - & 54.9 & 69.2 & 56.0/3.5 \\
            VideoLLaMA2-AoTD & \cmark & 56.0 & 71.1 & 57.2/3.5 \\
            Qwen2-VL-Instruct & - & 65.6 & 71.4 & 59.8/3.6 \\
            Qwen2-VL-AoTD & \cmark & 66.5 & 73.1 & 61.2/3.7 \\
            \bottomrule
        \end{tabular}
        \vspace{-5pt}
        \caption{Ablation results of CoT filtering and transferability.
        }
        \label{tab: ablation study}
        \vspace{-2em}
\end{table}

\subsection{Evaluation on Rationales}
\label{sec: rationale evaluation}
To verify whether the model has effectively learned multi-step reasoning through CoTs distillation, we analyze the rationales generated by the model. Specifically, we extract and evaluate the temporal and spatial information embedded within these rationales. This approach extends beyond merely assessing the correctness of the final answer, which could be influenced by biases or other external factors. By examining the reasoning process in detail, it enables a more accurate understanding of the model's ability to perceive and reason about spatial and temporal relationships.

\vspace{3pt} \noindent \textbf{Evaluation protocols.} 
We randomly select 200 samples from the STAR validation set and run inference on them using the suffix prompt, recording the generated rationales. From these rationales, we extract the predicted temporal windows and bounding boxes, comparing them to the ground truth. For the spatial part, we calculate the IoU between the predicted and ground truth bounding boxes. For the temporal part, we compute IoU and Recall, leveraging the frame-level annotations provided in the dataset.

\vspace{3pt} \noindent \textbf{Evaluation results.} 
Table \ref{tab: Temporal and Spatial ability evaluation} presents the evaluation results. For comparison, we also test UniVTG for temporal reasoning and OWL-ViT v2 for spatial reasoning. The results show that LNV-Instruct struggles to generate valid rationales, even when using the suffix prompt. In contrast, LNV-AoTD demonstrates comparable performance to specialized models in both spatial and temporal reasoning, indicating that the model successfully acquired these abilities through the distillation process.

\begin{table}[t]
    \centering
    \footnotesize
    % \small
    % \vspace{1em}
    \setlength{\tabcolsep}{8pt}
    \begin{tabular}{lccc}
        \toprule
        \multirow{2}{*}{\textbf{Model}} & \multicolumn{2}{c}{\textbf{Temporal Grounding}} & \textbf{Spatial Grounding} \\ 
         & \textbf{IoU (\%)} & \textbf{Recall (\%)} & \textbf{IoU (\%)} \\ \midrule
        UniVTG & 22.8 & 31.0 & - \\ 
        OWL-ViT v2 & - & - & 64.7 \\ 
        LNV-Instruct & \xmark & \xmark & \xmark \\ 
        LNV-AoTD & 21.7 & 34.0 & 45.2 \\ 
        \bottomrule
    \end{tabular}
    \vspace{-5pt}
    \caption{Temporal and spatial abilities evaluation results.}
    \label{tab: Temporal and Spatial ability evaluation}
    \vspace{-2em}
\end{table}

\section{Conclusion} \label{sec: conclusion}

We present Agent-of-Thoughts Distillation (AoTD), that aims to distill multi-step reasoning and spatial-temporal understanding into a large video-language model (Video-LLM). Our method introduces an agent-based system that automates the generation of Chain-of-Thoughts (CoTs) from various VideoQA datasets, by breaking down complex questions into manageable sub-tasks that can be addressed by specialized vision models. Extensive experiments validate that the distilled model significantly enhances performance on both MC-VQA and OE-VQA benchmarks, underscoring the effectiveness of our approach. We believe AoTD represents a promising future direction for advancing the reasoning abilities in Video-LLMs.

{
    \small
    \bibliographystyle{ieeenat_fullname}
    \bibliography{main}
}
% \clearpage
% \setcounter{page}{1}
% \maketitlesupplementary

\onecolumn
{
    \centering
    \Large
    \textbf{Unlocking Video-LLM via Agent-of-Thoughts Distillation}\\
    \vspace{0.5em}Appendix \\
    \vspace{1.0em}
}
\setcounter{page}{1}
\appendix
% {
%   \hypersetup{linkcolor=black}
%   \tableofcontents
% }
% \clearpage

\section{Limitation}
\label{sec: limitation}
Despite the advancements mentioned in the paper, several limitations remain and we leave them as future work: (i) similar to prior approaches~\cite{hu2024vpd,gao2024fact,fan2024videoagent}, the effectiveness of our agent-based system is contingent upon the progress of the underlying visual model components. Enhancing its ability to generalize across diverse datasets is essential for broader applicability;
(ii) while our primary focus has been on compositional VideoQA tasks~\cite{wu2021star}, and we have demonstrated improvements across a series of benchmarks, achieving holistic enhancements will require further exploration into creating a more balanced distribution of training data;
(iii) furthermore, our agent-based framework has the potential to address additional video-related tasks, such as video captioning and referring segmentation. 
We aim to expand our methodology to these domains, which could yield even more robust and versatile applications in the future.

\section{Experimental Details}
\label{sec: experimental details}

\subsection{Training Details}
For all models, their projection layers and language model are fine-tuned and visual encoder is frozen. We use a cosine learning rate schedule, with warm up ratio 0.03 and learning rate 4e-5. For both Instruct and AoTD setting, we fine-tune the model with batch size 48 and totally 1 epoch. We believe that longer training will get a better performance on in-domain benchmarks but maybe a destroy on out-of-domain benchmarks.

\subsection{Specialized Models Evaluation Details}
In this section we will show the details about each sub-task's evaluation from data preparation to evaluation metric.

\vspace{3pt} \noindent \textbf{Question decomposition.}
Since there may be multiple valid ways to decompose the same problem, we evaluate only the accuracy of the final output in this sub-task. Specifically, the model takes the query and instruction as input and generates an executable program. We replace all intermediate outputs within the program and focus on whether the final output matches the correct answer. If the decomposition is correct, the final output must align with the answer. Any programs that cannot be executed or that lead to an incorrect answer are considered failures.

\vspace{3pt} \noindent \textbf{Object detection.}
To evaluate the performance of detection models, we sample frames with scene graph annotations from the input video clip and provide them, along with the text query, as input to the model. The model then outputs a series of bounding boxes that exceed a confidence threshold. We select the bounding box with the highest confidence as the final output and calculate the IoU to assess accuracy.

\vspace{3pt} \noindent \textbf{Temporal grounding.}
Since scene graphs provide both the start and end frame IDs, as well as key frame IDs for each event, we use IoU and Recall as metrics to capture different aspects of model performance. The model takes the video clip and text query as input and outputs the predicted start and end frame IDs. We calculate IoU based on the alignment between the predicted and annotated start and end frame IDs, and we compute Recall using the key frame ID annotations to evaluate how well the model captures important frames.

\vspace{3pt} \noindent \textbf{Action recognition.}
For discriminative models, we provide the video clip and a list of action labels as input to complete a classification task. For generative models, we provide the video clip along with an instruction prompt, asking the model to generate five actions most relevant to the video, ranked by likelihood. We then use the top-ranked output from each model to calculate the Top-1 accuracy for both approaches.

\vspace{3pt} \noindent \textbf{Question answering.}
The evaluation of question answering follows a similar approach to previous methods. The model takes the video clip and question as input and returns an answer, from which we directly calculate the accuracy. The key difference between this sub-task and a standard QA task is that the answers are based on a series of information collected by preceding agents, allowing for a more accurate assessment of the model's pure question-answering ability.

\section{More Results} \label{sec: cot visualization}
Here, we introduce some examples to show the process from query to Chain-of-Thought using our agent-based system. We can find that our system is able to decompose complex questions into easier sub-tasks and the final CoT retains step-by-step problem-solving ideas and spatial-temporal information representing video understanding ability.

\begin{figure*}[h]
  \resizebox{\linewidth}{!}{
  \includegraphics{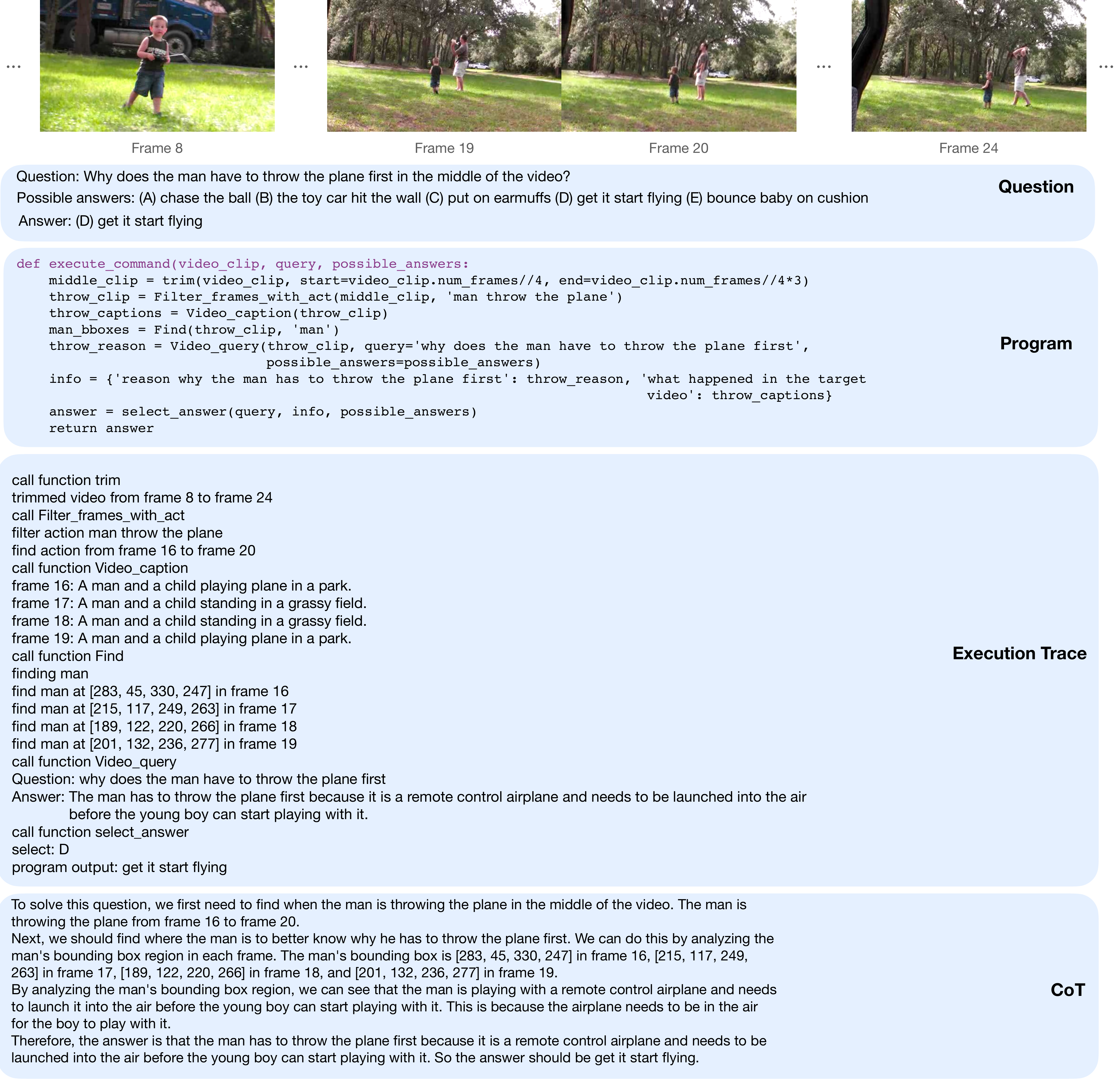}
  }
  \caption{
    Example form NExT-QA~\cite{xiao2021next}
  }  
\end{figure*}

\clearpage

\begin{figure*}[t]
  \resizebox{\linewidth}{!}{
  \includegraphics{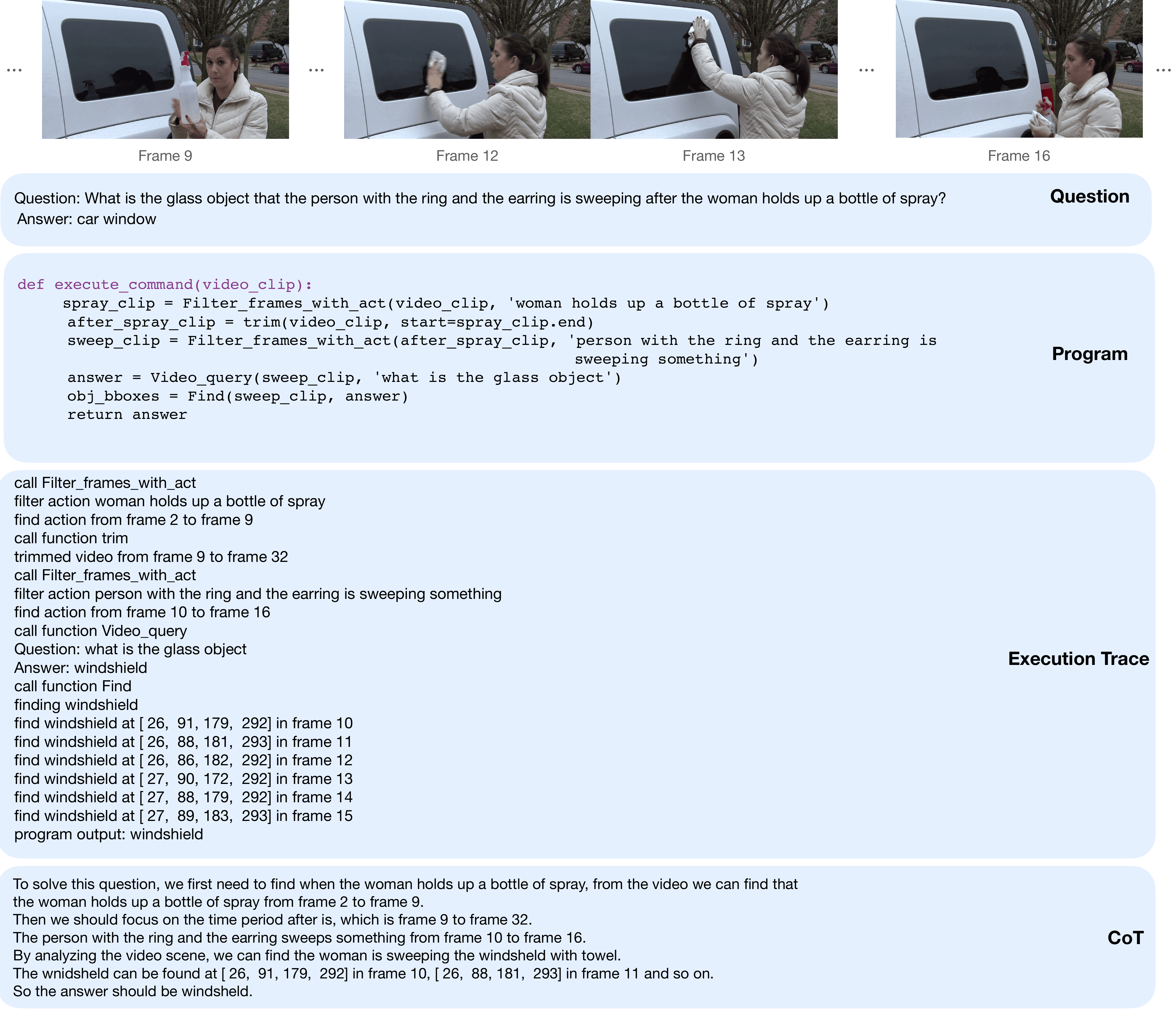}
  }
  \caption{
    Example form ANetQA~\cite{yu2023anetqa}
  }  
\end{figure*}

\section{Prompts} \label{sec: prompts}
In this section we present the prompts used in our agent-based system for generating programs, converting execution traces and filtering CoTs.

\subsection{Prompt for Program Generation}
For each video and query, we call a code LLM to decompose the query to a Python program under the guidance of the prompt below. We modify the ViperGPT \cite{suris2023vipergpt} prompt to adapt to the visual agents we use.
\begin{lstlisting}[language=Python]
def Query_Objs(clip, query):
    """
    Query the objects that appear in video clip and match the query descriptions.
    Parameters
    -------
    clip:
        a list of video frames.
    query:
        Description of the target object.
    Returns
    -------
    a list of bounding boxes of the objects that match the query.
    Examples
    -------
    #return white_objs
    def execute_command(video_clip):
        white_objs = Query_Objs(video_clip, "white object")
        return white_objs
    """

def Query_Actions(clip, obj=None):
    """
    Find the actions happened in the video clip, if obj is not None, query the actions related to it.
    Parameters
    -------
    clip:
        a list of the video frames.
    obj:
        object class which is used to query the actions related to it.
    Returns
    -------
    a list of actions classes happened in the video clip.
    Examples
    -------
    #return actions
    def execute_command(video_clip, query, possible_answers):
        actions = Query_Actions(video_clip)
        return actions
    """

def Filter_frames_with_act(clip, action):
    """
    filter a new video clip containing the time period in which the target action occurred
    Parameters
    -------
    clip:
        a list of video frames.
    action:
        the target action which is used to filter frames.
    Returns
    -------
    a new video clip ontaining the time period in which the target action occurred.
    Examples
    -------
    #return jump_clip
    def execute_command(video_clip, query, possible_answers):
        jump_clip = Filter_frames_with_act(video_clip, "person is jumping")
        return jump_clip
    """

def Filter_frames_with_obj(clip, obj):
    """
    filter a new video clip that the target object occured.
    Parameters
    -------
    clip:
        a list of video frames.
    obj:
        class or description about the target object.
    Returns
    -------
    a new video clip that the target object occured in it.
    Examples
    -------
    #return shoe_clip
    def execute_command(video_clip, query, possible_answers):
        shoe_clip = Filter_frames_with_obj(video_clip, "shoe")
        return shoe_clip
    """

def trim(clip, start=None, end=None):
    """
    Returns a new video clip containing a trimmed version of the original video at the [start, end] clip.
    Parameters
    ----------
    clip:
        a list of video frames.
    start : Union[int, None]
        An int describing the starting frame in this video clip with respect to the original video.
    end : Union[int, None]
        An int describing the ending frame in this video clip with respect to the original video.
    
    Returns
    -------
    a new video clip with start and end.
    """
def Find(clip, obj):
    """
    find all bounding boxes around a certain object in the video clip,
    and collates them into a collection of frames.
    Parameters
    ----------
    clip:
        a list of video frames.
    obj:
        the object to look for.
    Returns
    -------
    a new video clip composed of crops of the object.
    Examples
    --------
    # Return the shoe_clip
    def execute_command(video_clip, query, possible_answers):
        shoe_clip = Find(video_clip, "shoe")
        return shoe_clip
    """

def select_answer(query, info, possible_answers):
    """
    Uses a language model to choose the option that best answers the question given the input information.
    Parameters
    ----------
    query:
        the input question.
    info:
        Any useful information to answer the question.
    possible_answers:
        a list of possible answers to the question.
    Returns
    -------
    one answer chosen from the possible answers.
    Examples
    --------
    # Return the answer
    def execute_command(video_clip, query, possible_answers):
        clip_summary = Video_summary(video_clip)
        info = {
            "summary of the target video": clip_summary
        }
        answer = select_answer(query, info, possible_answers)
        return answer
    """
def exist(clip, query):
    """
    judge whether a object exists in the video.
    Parameters
    ----------
    clip:
        a list of video frames.
    query:
        query to the object class.
    Returns
    -------
    Return True if the object specified by query is found in the video, and False otherwise.
    Examples
    --------
    # Return the flag
    def execute_command(video_clip, query, possible_answers):
        flag = exist(video_clip, "shoe")
        return flag
    """
def Video_summary(clip, query):
    """
    give a brief summary of the video clip related to the query.
    Parameters
    ----------
    clip:
        a list of video frames.
    query:
        a question about the video.
    Returns
    -------
    return a brief summary of the video clip.
    Examples
    --------
    # Return the clip_summary
    def execute_command(video_clip, query, possible_answers):
        clip_summary = Video_summary(video_clip, query)
        return clip_summary
    """
Write a function using Python and the functions (above) that could be executed to provide an answer to the query. 

Consider the following guidelines:
- Use base Python (comparison, sorting) for basic logical operations, start/end, math, etc.
- Objects with mutiple names like "phone/camera", "cup/glass/bottle" with slash, input them as a whole object name.
- Just use the class and function appear above except for some base python operations.
- Only answer with a function starting def execute_command, do not answer any extra words and symbols before and after the function.
- No text that is not related to function can appear.
- the answer only begins with "def execute_command" and ends with "return answer".

Here are some examples of the function you should write:
-------
question: What else is the person able to do with the door?
possible answers: ["Hold the door.", "Put down the door.", "Close the door.", "Open the door."]
def execute_command(video_clip, query, possible_answers):
    door_clip = Filter_frames_with_obj(video_clip, "door")
    person_clip = Find(door_clip, "person")
    clip_summary = Video_summary(person_clip, query)
    door_actions = Query_Actions(person_clip, "door", possible_answers=possible_answers)
    door_actions = 
    info = {
        "actions the person able to do with the door else": door_actions,
        "summary of the target video": clip_summary
    }
    answer = select_answer(query, info, possible_answers)
    return answer
-------
Query: INSERT_QUERY_HERE
possible answers: INSERT_POSSIBLE_ANSWERS_HERE
\end{lstlisting}

\subsection{Prompt for Execution Trace Conversion}
After getting the execution trace by running the program step by step, we use a LLM to convert the trace into a natural language CoT. The LLM takes query, execution trace, possible answers (in MC-VQA) and execution trace as input. The instruction prompt is as follow:
\begin{lstlisting}[language=Python]
Given a video and a question, I wrote the function execute_command using Python, and the other functions above that could be executed to provide an answer to the query.
As shown in the code, the code will print execution traces.
I need you to rewrite the execution trace into a natural language rationale that leads to the answer.

Consider the following guidelines:
- Use all the bounding box information in the rationale, do not use words like "so on" to omit the bounding box, just write all of them into the rationale.
- Referencing the execution trace, write a reasoning chain that leads to the most common human answer. Notice that the output should be the same as the human answer, not necessarily the program output.
- If some part of the rationale lacks logic, add reasonable content to make it logical.


Here are some examples of the rantionale you should write:
-----
Question: What did the person do with the table?
def execute_command(video_clip, query, possible_answers, time_wait_between_lines, syntax):
    table_clip = Filter_frames_with_act(video_clip, 'person interacting with table')
    person_clip = Find(table_clip, 'person')
    table_bboxes = Find(table_clip, 'table')
    clip_summary = Video_summary(person_clip)
    person_action = Query_Actions(person_clip, 'table', possible_answers=possible_answers)
    info = {'actions the person do with the table': person_action, 'summary of the target video': clip_summary}
    answer = select_answer(query, info, possible_answers)
    return answer
Execution trace:
call Filter_frames_with_act
filter action person interacting with table
find action from frame 2 to frame 11
call function Find
finding person
find person at [139, 141, 229, 342] in frame 2
find person at [151, 123, 242, 349] in frame 3
find person at [153, 121, 242, 274] in frame 4
find person at [158, 123, 255, 261] in frame 5
find person at [163, 124, 270, 262] in frame 6
find person at [153, 121, 242, 351] in frame 7
find person at [95, 113, 196, 316] in frame 8
find person at [83, 113, 196, 285] in frame 9
find person at [112, 116, 201, 332] in frame 10
call function Find
finding table
find table at [183, 140, 269, 257] in frame 2
find table at [194, 131, 269, 255] in frame 3
find table at [227, 129, 269, 252] in frame 4
find table at [226, 165, 269, 258] in frame 5
find table at [233, 170, 270, 259] in frame 6
find table at [217, 129, 269, 256] in frame 7
find table at [217, 122, 270, 254] in frame 8
find table at [221, 123, 269, 256] in frame 9
find table at [225, 125, 270, 263] in frame 10
call function Video_summary
summary result: The video shows a man in a kitchen, bending over and holding an orange object, surrounded by various kitchen items and furniture, with a focus on his actions and the domestic setting.
call function Query_Actions
Query table
Answer: tidied up.
call function select_answer
the information used: - actions the person do with the table: tidied up.
- summary of the target video: The video shows a man in a kitchen, bending over and holding an orange object, surrounded by various kitchen items and furniture, with a focus on his actions and the domestic setting.
program output: Tidied up.
Rationale:
To solve this question, we first have to find when did the person interact with the table.
From the video, we can see that the person is interacting with the table from frame 2 to frame 11.
In this time period, we can find person at [139, 141, 229, 342] in frame 2, [151, 123, 242, 349] in frame 3, [153, 121, 242, 274] in frame 4 and so on.
Table can also be found at [183, 140, 269, 257] in frame 2, [194, 131, 269, 255] in frame 3, [227, 129, 269, 252] in frame 4 and so on.
By analyzing the person and table bounding box region, we can see that the person is holding an orange object to clean the table in the kirchen environment.
So the answer should be tidied up.
------------------------------------------------
Now, look the question, program and execution trace, please transfer these information to a rantionale.
Question: INSERT_QUESTION_HERE
INSERT_PROGRAM_HERE
Execution trace:
INSERT_EXECUTION_TRACE_HERE
Rationale:
\end{lstlisting}

\subsection{Prompt for CoT Filtering}
In order to obtain high quality distillation data, we continue using LLM to filter CoTs. We prompt the LLM to select those CoTs that are truly helpful for solving questions and reflect the step-by-step thinking process. The prompt is as follows:
\begin{lstlisting}[language=Python]
I will give you a question and a rationale to solve the question, you need to judge whether the rationale is thinking step by step and helpful to solve the question.
If yes, return True, If not, return False. no need to explain.
Here is the question and rationale:
Question: INSERT_QUESTION_HERE
Rationale: INSERT_RATIONALE_HERE
\end{lstlisting}

\subsection{Prompt for Inference}

\begin{tcolorbox}
Question: question content \\
Answer in one word or phrase. / Explain the rationale to answer the question.
\end{tcolorbox}

\begin{tcolorbox}
Question: question content \\
Options: \\
(A) option content \\
(B) option content \\
(C) option content \\
(D) option content \\
Answer with the option's letter from the given choices directly and only give the best option. / Explain the rationale to answer the question.
\end{tcolorbox}


\end{document}